\documentclass[conference]{IEEEtran}
\IEEEoverridecommandlockouts
\usepackage{amsmath,amssymb,amsfonts}
\usepackage{algorithmic}
\usepackage{graphicx}
\usepackage[inline, shortlabels]{enumitem}
\usepackage{tabularx}
\usepackage{caption}
\usepackage[T2A,T1]{fontenc}
\usepackage[english]{babel}
\usepackage{ragged2e}
\usepackage{hyperref}
\usepackage{pifont}
\usepackage{footmisc}
\usepackage{pdfpages}
\usepackage{booktabs}
\usepackage{csquotes}
\usepackage{smartdiagram}
\usepackage[inkscapeformat=png]{svg}
\usepackage{textcomp}
\usepackage{xcolor}
\def\BibTeX{{\rm B\kern-.05em{\sc i\kern-.025em b}\kern-.08em
    T\kern-.1667em\lower.7ex\hbox{E}\kern-.125emX}}
\usepackage{cite}
\usepackage{amsmath}
\newcommand{\probP}{\text{I\kern-0.15em P}}

\usepackage{etoolbox}
\patchcmd{\thebibliography}{\section*{\refname}}{}{}{}

\makeatletter
\newcommand{\linebreakand}{%
  \end{@IEEEauthorhalign}
  \hfill\mbox{}\par
  \mbox{}\hfill\begin{@IEEEauthorhalign}
}
\makeatother

\begin{document}

\title{Towards a Multi-Agent Simulation of Cyber-attackers and Cyber-defenders Battles\\
}


\author{

\IEEEauthorblockN{Julien Soulé}
\IEEEauthorblockA{\textit{Thales Land and Air Systems, BL IAS}}
\IEEEauthorblockA{\textit{Univ. Grenoble Alpes,} \\
\textit{Grenoble INP, LCIS, 26000,}\\
Valence, France \\
julien.soule@lcis.grenoble-inp.fr}

\and

\IEEEauthorblockN{Jean-Paul Jamont\IEEEauthorrefmark{1}, Michel Occello\IEEEauthorrefmark{2}}
\IEEEauthorblockA{\textit{Univ. Grenoble Alpes,} \\
\textit{Grenoble INP, LCIS, 26000,}\\
Valence, France \\
\{\IEEEauthorrefmark{1}jean-paul.jamont,\IEEEauthorrefmark{2}michel.occello\}@lcis.grenoble-inp.fr
}



\linebreakand

\hspace{-5ex}
\IEEEauthorblockN{Paul Théron}
\hspace{-5ex}
\IEEEauthorblockA{
\hspace{-5ex}
\textit{AICA IWG} \\
\hspace{-5ex}
La Guillermie, France \\
\hspace{-5ex}
paul.theron@orange.fr}

\and

\hspace{2ex}

\IEEEauthorblockN{Louis-Marie Traonouez}
\IEEEauthorblockA{\textit{Thales Land and Air Systems, BL IAS} \\
Rennes, France \\
louis-marie.traonouez@thalesgroup.com}}

\maketitle

\begin{abstract}

As cyber-attacks show to be more and more complex and coordinated, cyber-defenders strategy through multi-agent approaches could be key to tackle against cyber-attacks as close as entry points in a networked system.
This paper presents a Markovian modeling and implementation through a simulator of fighting cyber-attacker agents and cyber-defender agents deployed on host network nodes. It aims to provide an experimental framework to implement realistically based coordinated cyber-attack scenarios while assessing cyber-defenders dynamic organizations. We abstracted network nodes by sets of properties including agents' ones. Actions applied by agents model how the network reacts depending in a given state and what properties are to change. Collective choice of the actions brings the whole environment closer or farther from respective cyber-attackers and cyber-defenders goals.
Using the simulator, we implemented a realistically inspired scenario with several behavior implementation approaches for cyber-defenders and cyber-attackers.

\end{abstract}

\begin{IEEEkeywords}
cyber-defense, multi-agent system, simulation, Dec-POMDP
\end{IEEEkeywords}

\section{Introduction}



\noindent
Internet of Things development has highlighted an increase of the attack surface in networked systems offering attackers more ways to infiltrate.
Considering this context, the \textquote{AICA IWG}\footnote{This working group (see \url{https://www.aica-iwg.org/}) succeeded the NATO \textit{Research Task Group IST-152} which focused on the concept of \textquote{Intelligent agents , Autonomous and Trusted for Cyber Defense and Resilience}.} continued the development of the \textquote{Autonomous Intelligent Cyber-defence Agent} (AICA).
The AICA must be able to be autonomously deployed on a host system to detect, identify and characterize anomalies/attacks, develop and manage the execution of countermeasures and dialogue with the outside. To this end, it is designed as proactive, stealthy and capable of learning.

Furthermore, considering cyber-defenders and cyber-attackers deployed in a dynamic networked infrastructure, issues related to cyber-defense collective strategies of action are making up challenges as for modeling and simulation tools.

\noindent
The main contribution is a simulation model whose main interest is to provide a common framework to implement and evaluate types of cyber-defenders agents and cyber-attackers agents on the same networked environments. Integrating existing attack scenarios into cyber attackers aims to assess the effectiveness of agents, analyze or visualize their various behaviors, to train cyber-defense agents against cyber-attack agents, etc. Additionally, it allows spotting causes of malfunctions or poor performance and improving the cyber-defender agents. While it could raise several AI-related challenges such as the network environment generation, the paper particularly focuses on multi-agent learning and collective decision-making.

Section II gives an overview of relevant related works addressing aspects of modeling cyber-attackers and cyber-defender agents in networked host systems. Section III introduces the abstract modeling for the simulation, its technical implementation and its use for integrating and evaluating attack/defense scenarios. In section IV, we present a MITRE ATT\&CK~\cite{MITREATTACKWebiste} based case study and its full implementation in the simulator as an attack/defense scenario for cyber-attackers and cyber-defenders. It also presents results of the scenario executions to assess the simulation relevancy. Section V, concludes on limitations to overcome and future perspectives.


\section{Modeling related works}

\noindent

Few works directly address the modeling of cyber-attackers and cyber-defenders fighting in a networked host system. Indeed, available related works mostly provide a way to model attack actions for a single cyber-attacker focusing on specific attack scenarios while optional cyber-defense is mostly thought in reaction.


\

\noindent
\textbf{Attack graphs}: \quad Attack graphs~\cite{CPhilips1998} are graphical representations of the different ways an attacker can exploit vulnerabilities in a networked system. They represent the system as a set of nodes (such as computers, applications, or network connections) and the possible attacks as edges between those nodes. The graph shows how an attacker can move from one node to another by exploiting vulnerabilities and express the consequences on the network~\cite{CPhilips1998}.
Attack graphs can be used to identify the most critical vulnerabilities in a networked system and to help the defender prioritize their efforts to secure those vulnerabilities in that system.


\noindent
\textbf{Attack-Defense trees}: \quad Attack-Defense trees~\cite{BKordy2010} (AD trees) are graphical models representing the attacker's goals and the defender's countermeasures as a tree structure. AD trees provide a more abstract representation of the system and the attackers goals, while attack graphs provide a more concrete representation of the system's components and their relationships. The root of the tree represents the cyber-attackers' ultimate goal. The associated sub-nodes of the branches represent different attack strategies that the attacker might use to achieve their goal. They can be decorated with preventive or reactive defender's countermeasures (firewalls, intrusion detection systems, incident response plans\dots).
AD trees allow identifying the weakest points in a system's defense~\cite{BKordy2010}.

\noindent
\textbf{Petri nets modeling}: \quad As Petri nets can be used to describe concurrent processes, some works have been pushing modeling attackers and defenders in a networked system.
Extracted attacks from databases can be modeled with Petri nets to integrate cyber-attackers and cyber-defenders, their strategies, and the cost of their actions as in ~\cite{MPetty2022}. Petri nets also show to be used to model structured query language injection attacks to include players' strategies~\cite{JBland2020}.
They are used as a framework for assessing and comparison between several attack models.
In ~\cite{SYamaguchi2020}, IoT malware concerns have also been addressed for \textquote{Mirai} malware using a \textquote{white worm} solution expressed as a formal model with extended agent-oriented Petri nets. This modeling allows simulating an example of battle between the white-hat worm and the Mirai malware.

\noindent
\textbf{Game models}: \quad Some works have proposed modeling interactions of attackers or defenders in a network as players in a game, where each player has a set of actions that they can take.
Some notable works include: Panfili et al.~\cite{MPanfili2018} where an attacker vs. defender multi-agent general sum game is used to find an optimal trade-off between prevention actions and costs; Attiah et al.~\cite{AAttiah2018} where a proposed dynamic game theoretical framework is to analyze the interactions between the attacker and the defender as a non-cooperative security game; and Xiaolin et al.~\cite{CXiaolin2008} using Markov process models to assess risks in networked systems.

\noindent
Some game-theoretical approaches fall in the \textquote{Partially Observable Stochastic Game} (POSG) framework or more specifically in \textquote{Decentralized Partially Observable Markov Decision Process} (Dec-POMDP). Both POSGs and Dec-POMDPs are frameworks for mathematical modeling of decision-making problems in which agents interact with each other and in a stochastic environment~\cite{beynier2010}. In a POSG, a group of agents interacts with a stochastic and partially observable environment. Each agent act according to his own observations and one local policy. Agents may have different goals as each agent has its own reward function and the game is generally assumed to be non-cooperative~\cite{jk2020}. In a Dec-POMDP, several agents can have a common reward function and can coordinate their actions to achieve a common goal, especially by being able to communicate~\cite{bernstein2013}.

\



\noindent
In order to define a modeling, we took a use case from the AICA~\cite{theron_autonomous_2021}. We are interested in modeling a network environment made up of \textit{nodes} on which cyber-attackers and cyber-defenders \textit{agents} can be deployed to observe and act. These nodes can be described by a set of \textit{properties} related to processes, file systems, operating systems, hardware architecture, etc.
The \textit{observations} and \textit{actions} of the agents are conditioned by their own properties (including the properties known by them) and uncertainties. For example, reading a given file or remapping ports may require an elevated privilege level; or the reception of data from a physical sensor is not ensured at all times.
Each agent applying actions modifies the properties of one or more nodes. This changes the state of the environment, making the agents closer or farther from their objective.
The key features of this description are encompassing notions like uncertainty in observations, conditions in actions for state transitioning and metrics that we consider to be expressed within a Dec-POMDP modeling.

\section{Simulation Model}

\subsection{General Dec-POMDP modeling of environment and agents}


\begin{figure*}[]
    \centering
    \includegraphics[width=0.97\textwidth]{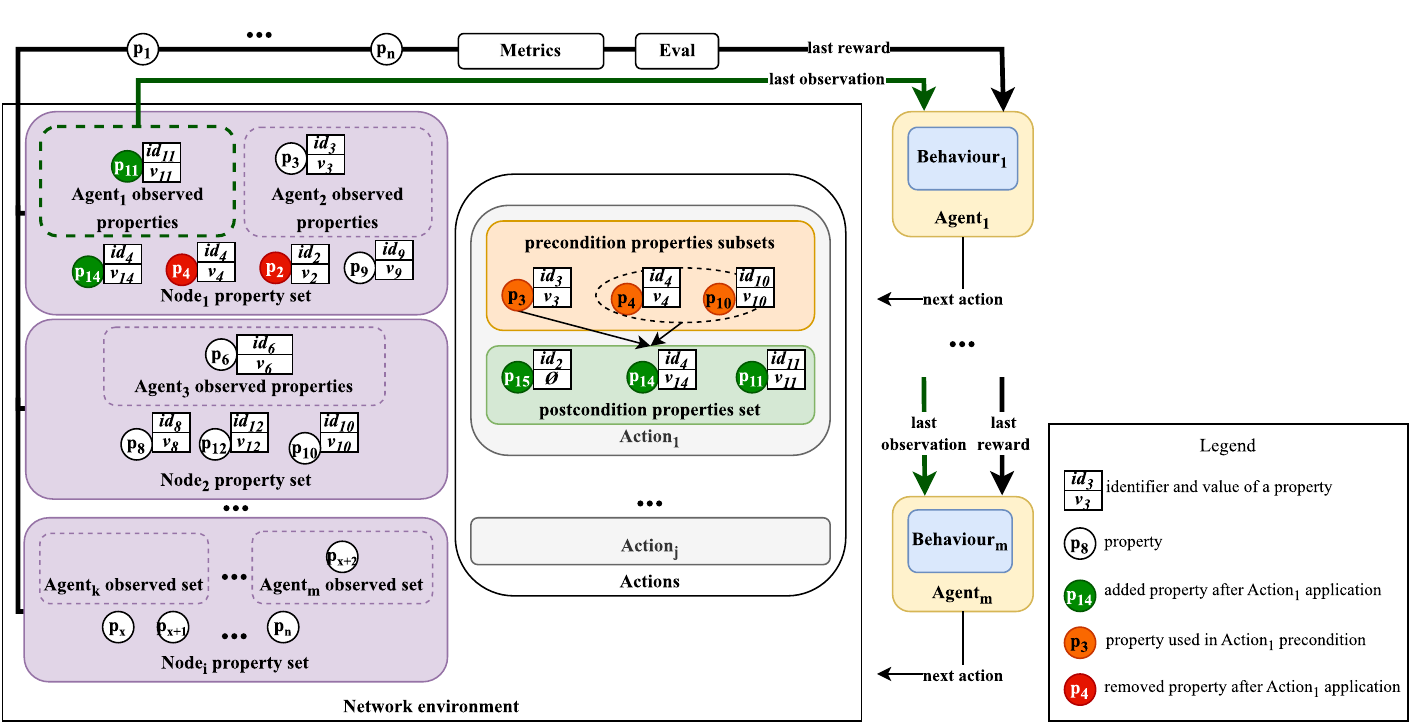}
    \caption{An illustrative view of the simulation model}
    \label{fig:model_example_illustration}
\end{figure*}

\noindent
From a global view the proposed Dec-POMDP model expresses an environment state as the set of nodes properties including agents observable properties. We define a property as a couple made of an identifier and a value. The environment state is changed when an action is applied by an agent. An action can be applied only if the boolean property-based pre-condition is satisfied in the current environment state. The resulting state is then modified depending on post-condition ultimately leading some new properties to be added while some others are deleted. After an action is successfully applied by an agent, observable properties of the same agent are returned to it as observations from that new state. A reward is also computed based on the current state and returned to the agent. An agent is chosen to be modeled as a behavior function which has to select the next action to be made depending on received observations and rewards.

There are different ways for several agents to be executed in a same environment tweaking with the number of agents to execute in a time step and the number of actions to be played by an agent in a time step. Even though not realistic, we chose the \textquote{Agent Environment Cycle}~\cite{jk2020} as a first approximation by having several agents playing one action in each turn in a sequential cyclic manner. The iteration cycle is presented through an illustrative informal view of the simulation model in Figure~\ref{fig:model_example_illustration}. It shows $i$ nodes with their properties including the $m$ agents' observed ones; and how each of the available $j$ actions associates a pre-condition set of properties subsets to a subset of new properties to be added in the environment optionally deleting obsolete properties having the same identifiers as the new properties' ones: \begin{enumerate*}[label=\arabic*),itemjoin={;\quad}]     
    \item An agent chooses an action from previous observations and rewards according to a behavior function. In Figure~\ref{fig:model_example_illustration}, as $Agent_1$ begins its first turn, it receives only initial observations ($p_{1}$) and zero rewards and chooses $Action_1$
    
    \item The environment is updated by a transition function depending on the current state and the action taken by the agent (change of properties once the pre-condition is satisfied). An action is used to change the environment properties by updating the relation between property identifiers and property values.
    For instance, in Figure~\ref{fig:model_example_illustration}, in current state, $Node_1$ properties are $p_1,p_3,p_4,p_2,p_9$. When $Action_1$ is applied, the relation associate subsets $\{p_3\}$ or $\{p_4, \allowbreak p_{10}\}$ to $\{p_{15}, \allowbreak p_{14}, \allowbreak p_{11}\}$. The property based pre-condition can be understood as $p_3 \lor (p_4 \land p_{10})$. As $p_{15}$ and $p_{14}$ are identified by $ID_4$ and $ID_2$ which already respectively define $p_{2}$ and $p_{4}$, $p_{4}$ and $p_{2}$ are deleted and $p_{11}$ and $p_{14}$ are added ($p_{15}$ is not added as $id_2$ is not associated with any value)
    
    \item Observed properties are returned to the current executor agent for its next turn. In Figure~\ref{fig:model_example_illustration} $p_{11}$ and $p_1$ are returned after $Action_1$ is applied.

\end{enumerate*}

\noindent
Agents are selected following a sequential order. Each one receives last observations and rewards from their last turn (or just initial observation and zero rewards if they are on their first turn); chooses the next action to play in its current turn. Once the last agent has finished playing (such as $Agent_m$), rewards are computed and sent to cyber-attackers and cyber-defenders based on the evaluation of the collected metrics from last state. Then, the agents play again following the same sequential order for another iteration.

\subsection{Formal Dec-POMDP modeling}

We set the elements related to the properties of the nodes, agents and actions of the following environment:

\begin{itemize}

    \item $Ag = \{ag_1,..,ag_{|Ag|}\}$: The set of agents (cyber-attackers and cyber-defenders).

    \item We call the couple $p = (id_{j}, v_{j})$ with $id_j \in {ID}$ and $v_j \in V$, a property.
    \begin{itemize}
        \item $ID$: The set of property identifiers optionally indicating how are organized the properties in a non-flat data structure (such as $PC1.processes.agents.agent1$). These property identifiers may be used for a file path, the type of operating system used in a node, a used command line by an agent\dots
        \item $V$: The set of property values. These may include the content of a file, a full description of the operating system, the output result of a command line\dots
    \end{itemize}

    \item $P_{j} = \{ p_1, .., p_{|P_{j}|} \}$: The set of the $p_{l}$ properties (with $l \in \{1,..,|P_{j}|\}$) of node $j$ ($j \in \mathbb{N} $). For example, such properties may include some running process IDs, files list in a folder, type of operating system with description, specific knowledge of an agent, etc.
    \begin{itemize}
        \item $P = P_1 \cup P_2 .. \cup P_{|P|} $: The set of all the node properties.
    \end{itemize}

    \item $Obs: \mathcal{P}(P) \times Ag \rightarrow \mathcal{P}(P_{Ag}), P_{Ag} \subset P$: A relation which associates node properties and an agent with the observed property subset by the agent.
    
    \item $Action: P_{pre} \rightarrow P_{post}$: A relation which associates a property subset implied by an equivalent conjunctive boolean pre-condition ($P_{pre} \subset \mathcal{P}(P)$) to a subset of all of the properties of the post-condition ($P_{post} \in \mathcal{P}(P)$). For example, the properties $p_1 = (agent\_X\_privilege\_level, \allowbreak root)$, $p_2 = (agent\_X\_accessed\_text\_editor, \allowbreak Vim)$ and $p_3 = (agent\_X\_bashrc\_known\_filepath, \allowbreak /home/user/.bashrc)$ can make a pre-condition ($p_1 \land p_2 \land p_3$) to associate a new set of property containing $p4 = (bashrc\_file\_modified\_by\_X\_agent, \top)$. Two pre-condition subsets can be associated to the same post-condition subset to model a boolean disjunction.

    \item $Metrics: \mathcal{P}(P) \times A \rightarrow \mathbb{R}^{n}$: Gives metrics associated with a set of properties and joint action. For example, the number of nodes still active, lateral moves, etc.

\end{itemize}

Using the formal description of a Dec-POMDP~\cite{OliehoekA16}, we propose the following model:

\begin{itemize}
    \item $S = \{s_1, ..s_{|S|}\}, s_{i} \subseteq P \: and \: 1 \le i \le |S|$: The space of states as possible property sets.

    \item $A_{i} = \{a_{i}^{1},..,a_{i}^{|A_{i}|}\}, a_{i}^j \in Action \: and \: 1 \le j \le |A_i|$: The set of possible actions for agent $i$.

    \item $T$ : The set of conditional transition probabilities between states
    \begin{itemize}
        \item With $T(s,a,s') = \probP(s'|s,a)$, the relation which associates probability to go to state $s' \in S$ from state $s \in S$ knowing we played $a = (P^a_{pre} \times P^a_{post}) \in A$ with $P^a_{pre} \subset \mathcal{P}(P)$ and $P^a_{post} \in \mathcal{P}(P)$
        \item With $\probP(s'|s,a) = 0$ if $s$ does not satisfy the pre-condition of $a$ (i.e $\exists \: P_{pre_s}^{a} \in P_{pre}^{a} \: | \: P_{pre_s}^{a} \not\in \mathcal{P}(s)$).
        \item With $s' = (s - \{p_l=(id_l, v_l) \: | \: p_l \in s \: and$ $id_l \in \{id_k \: | \: (id_k, v_k) \in P^a_{post} \: and \: v_k \neq \varnothing\}\}) \cup P^a_{post}$
    \end{itemize}
    
    \item $R: S \times A \rightarrow \mathbb{R}^2 = Eval \circ Metrics$: The reward function that takes a state and an action and associates a performance indicator (using the state's metrics) for attackers and defenders.
    \begin{itemize}
        \item With $Eval: \mathbb{R}^{n} \rightarrow \mathbb{R}^2$, associates a metric vector to a a reward for cyber-attackers and cyber-defenders.
    \end{itemize}
    
    \item $\Omega_{i} \subset Range(Obs \: | \: \{ (s, ag_i) | s \in S \: and \: ag_i \in Ag \}) \subset P$: The set of observable properties for agent $ag_i$. For example, the content of a file, the log output of a command, the result of a port scan, etc.
    \begin{itemize}
        \item $\Omega = \Omega_1 \cup \Omega_2 .. \cup \Omega_{|Ag|} = Range(Obs)$: The set of all the observable properties for all agent.
    \end{itemize}

    \item $O$ : The set of conditional observation probabilities.
    \begin{itemize}
        \item With $O(s',a,o) = \probP(o|s',a)$, the relation which associates the probability to observe an observation $o \subset \Omega$ from state $s' \in S$ induced by $a \in A$
        \item With $\probP(o|s',a) = 0$ if the state $s' \in S$ does not contain the properties of $o \subset \Omega$ (i.e $o \not\in \mathcal{P}(s')$). For example, an agent plays the action $x\_reads\_a\_log\_file$, a new state results from which a property belonging to the knowledge of agent x is $(log\_file\_content\_known\_by\_x, \allowbreak abc)$. This property will be therefore included in the returned observations to agent x. 
    \end{itemize}

\end{itemize}

\subsection{Attack/defense scenarios integration\label{sec:ad_integration}}

\noindent
From a raw perspective, the proposed formal Dec-POMDP modeling relies on actions to simulate how a real networked system would react including vulnerabilities and countermeasures applied by cyber-attacker and cyber-defender agents.

A first challenge is to build a representative attack/defense scenario of a networked system comprising vulnerabilities to allow rendering an attack by linking the only available pieces of information (such as known tactics, techniques and procedures from MITRE ATT\&CK) and by choosing relevant defense countermeasures (from MITRE ATT\&CK mitigations) and a deployment environment. A second challenge is to establish the actions to match the attack/defense scenario. As actions modify the environment properties, they also impact possible states space and the transitions between them.
Moreover, when considering a low abstraction level, numerous simple actions may allow describing the operated changes in the network finely. Yet, doing so increases the number of actions, and even more the number of states for they are combinations of action effects.

These challenges are directly linked to studied issues about automated generation of attack graphs using available databases optionally integrating artificial intelligence techniques as in ~\cite{GFalco2018}. We do not intend to focus more on these issues as they are out of the scope of this work.

\

\noindent
\textbf{MITRE ATT\&CK integration approach}: We suggest a high-level manual approach we used to integrate MITRE ATT\&CK information as an AD tree for it formalizes actions to be played in a scenario and their interactions with the environment. It is aimed at being helpful to establish the attack/defense actions to be finally integrated in the simulator:
\begin{enumerate*}[label=\arabic*),itemjoin={;\quad}]

    \item For a given Advanced Persistent Threat (APT), we identified relevant tactics and techniques and procedures from MITRE ATT\&CK that seemed relevant for a networked system
    
    \item We produced a description linking identified tactics together and associated techniques, sub-techniques and procedures to create a scenario that describes how the APT group could attack the networked system. This step defines the network topology with its main properties

    \item We created an AD tree as proposed in ~\cite{BKordy2010} with tactics as top action goals while techniques, sub-techniques and procedures are in the lower part of the tree. We made sure to have several paths to reach a same top-action goal. We paid attention to define each attack action with property based pre-condition and property post-conditions in the environment

    \item We extracted the MITRE ATT\&CK techniques/sub-techniques related detection and mitigations we added in the AD tree to decorate the attack nodes. We paid attention to define each defense actions with property based pre-condition and property post-conditions in the environment.


\end{enumerate*}

\subsection{Simulation model implementation}

\noindent
Potential works to implement our model include: NeSSi2~\cite{DGrunewald2011} which is an agent-based simulation platform aiming to model only packet-level description of a networked system and the effects of DDoS attacks; and Kotenko et al.~\cite{IKotenko2007} which relies on OMNet++~\cite{Varga2010} to model and simulate cooperative cyber-defense agents against network attacks combining discrete-event simulation, multi-agent approach and packet-level simulation of network protocols.
However, among these, none can fully meet both the consideration of a multi-agent cyber environment for a Dec-POMDP model.

\begin{figure}
    \centering
    \includegraphics[width=\linewidth]{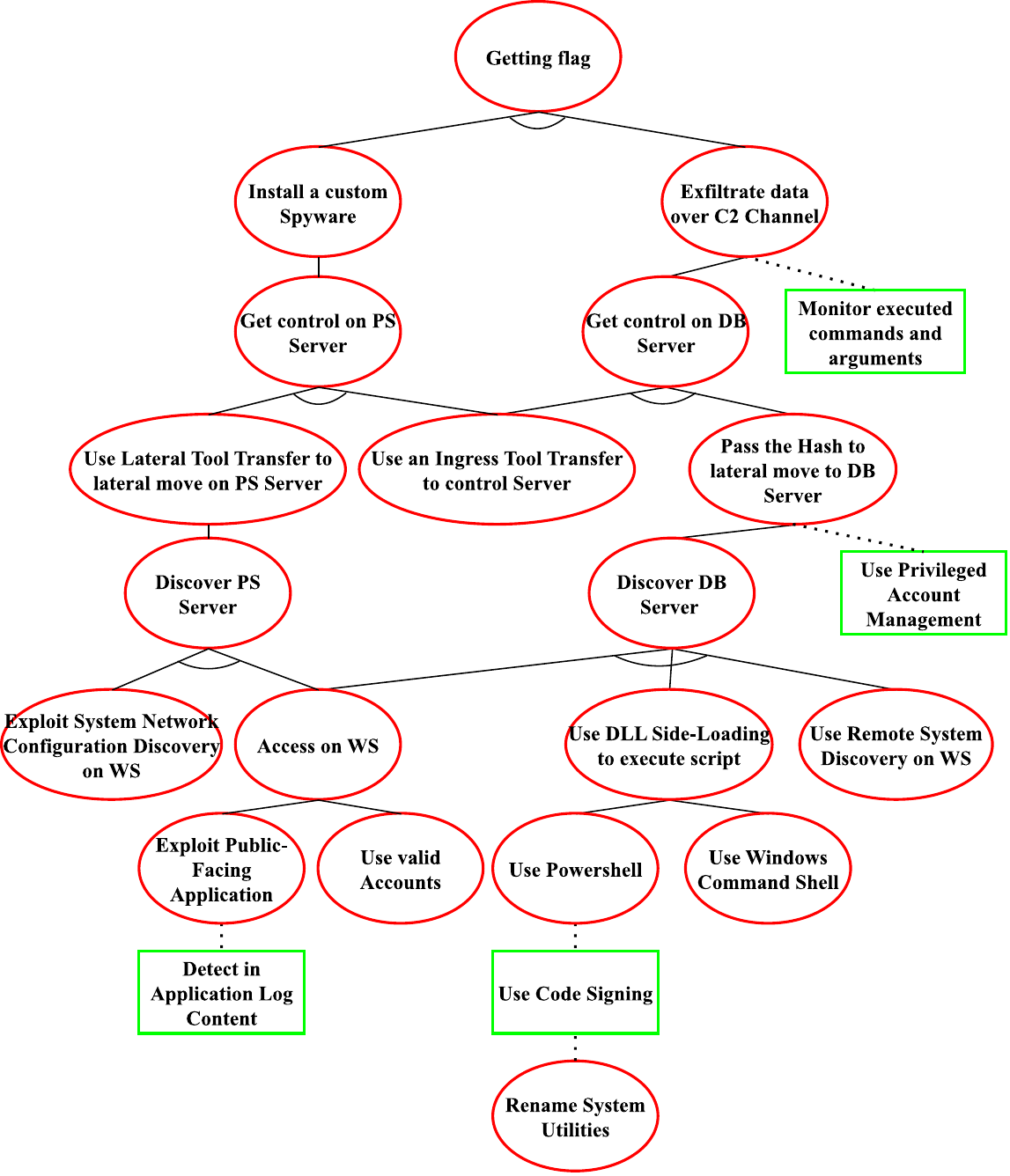}
    \caption{An overview of the proposed attack/defense AD Tree}
    \label{fig:ADTree}
\end{figure}

Yet, we identified discrete-event simulators with a single cyber-attacker such as CYST\cite{drasar_session-level_2020} or CyberBattleSim~\cite{cyberbattlesim}, which both provide a suited network simulation and evaluation approach towards an implementation of our model as their underlying models can be extended for several agents. Inspired by these approaches, we used \textit{PettingZoo}~\cite{jk2020} as a fundamental platform to implement our Dec-POMDP model onto which we aimed to implement a simulated network. \textit{PettingZoo} provides a framework where the designer has tools to facilitate the implementation of the space of observations, actions, management of agents at each turn and associated rewards.


The development of our model has led to the \textquote{Multi Cyber Agent Simulator} (MCAS)~\cite{MCASWebsite} simulator. In the current state of development, this simulator allows loading/saving a \textit{json} file describing the nodes properties and actions of the environment and the defined agents with their behaviors; and launching the execution of the agents of this environment in turn-by-turn mode via the terminal. It is possible to view the environment properties in real time and visualize the environment in the form of a graph. Metrics are displayed as well.

\section{MITRE ATT\&CK based case study}


\noindent
Intending to assess coordinated attacks and collective defense, we chose GALLIUM APT (a cyberespionage group active since 2012), for it allows defining several concurrent attacks.

\subsection{Network topology}

\begin{figure}
    \centering
    \includegraphics[width=\linewidth]{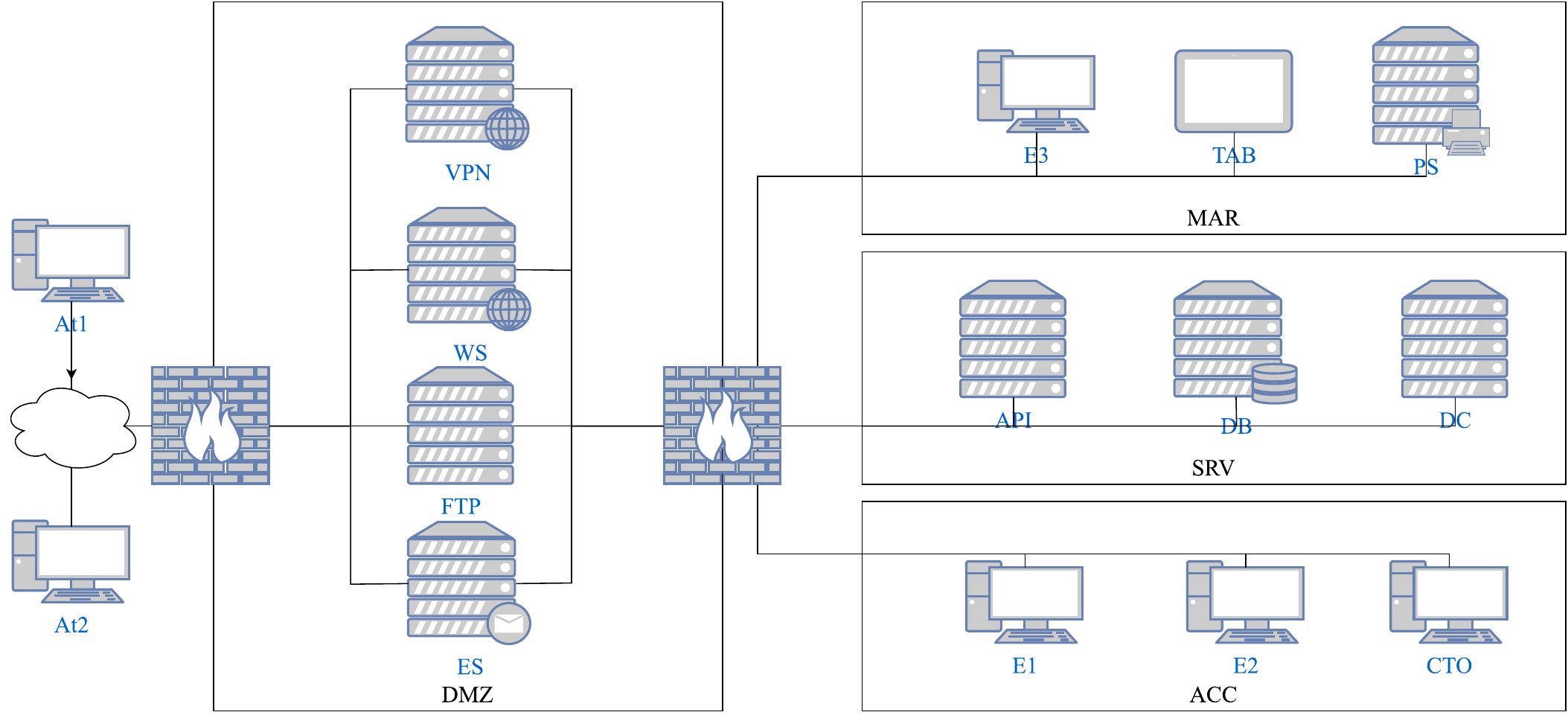}
    \caption{Proposed small-scale company network topology}
    \label{fig:scenario_network_topology}
\end{figure}

\noindent
Based on some GALLIUM APT tactics we selected some associated techniques/sub-techniques to propose a small company like networked environment, presented in Figure~\ref{fig:scenario_network_topology}. It is divided in 5 subnets communicating through implicit routers positioned after a firewall:
The outside subnet used to represent external attackers as if in the same network for convenience. It includes two desktop computers (At1 and At2).
The Demilitarized Zone (DMZ) subnet is used to separate devices that are accessible from the Internet from the rest of the company's network. The servers in the DMZ include a web server (WS), an email server (ES), a VPN server (VPN), and a FTP server (FTP), all connected bidirectionally both to outside and inside the company network.
The first local area network (ACC) subnet used to connect devices within the accounting department of the company where employee are working in. It contains two employee workstations (E1 and E2) and a Chief Technical Officer workstation (CTO), all connected bidirectionally to the DMZ. 
The second local area network (MAR) subnet used to connect devices within the marketing department of the company. It contains a printer server (PS), one employee workstations (E3) and tablet terminal (TAB) connected via a wireless access point, all connected bidirectionally to the DMZ. 
The third local area network (SRV) subnet used to connect the company devices providing services. It contains an API server (API), a database server (DB), and a domain controller (DC).

\subsection{Scenario and agent implementation with evaluation}

\begin{figure}
    \centering
    \includegraphics[width=\linewidth]{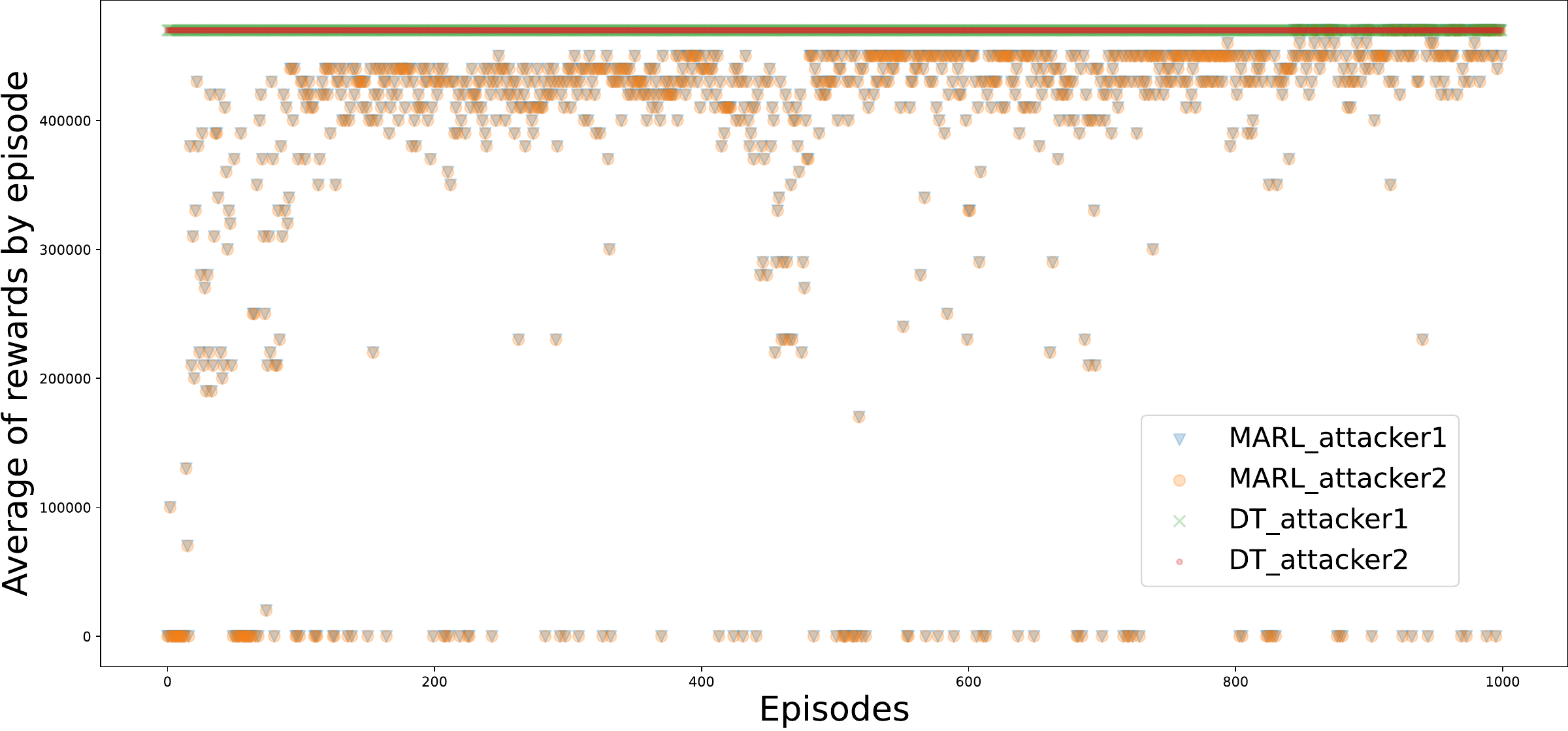}
    \caption{An evolution of the rewards average according to episodes in small-scale tests with MARL and Decision Tree Approaches with inactive cyber-defense
    }
    \label{fig:graphs}
\end{figure}

\noindent
The cyber-attacker agents are initially deployed on At1 and At2 and the cyber-defender agents are deployed on WS and DB. The ultimate attackers' goals is to get data from the DB server and installing one spyware on the printer server PS. Following our approach in ~\ref{sec:ad_integration} we propose an AD tree presented in Figure~\ref{fig:ADTree}. It only shows the paths of the attacks to follow to reach the ultimate goal while the defender actions can prevent these at several stages of the attack.
We first interested in setting up the two cyber-attackers behaviors, then the two cyber-defenders' ones. We simulated an abstracted version of the scenario over 1000 episodes.



\noindent
\textbf{Random approach}: \quad The random agent only choose its actions by exploring the whole action space without any criteria until reaching the goal. In our case study, the shortest action path for attackers to reach the ultimate goal contains 16 different actions among the 30 defined actions, hence a low probability of $(1/30)^{16}$.
This approach allows getting a benchmark of unexpected edge failure cases and to compare with other types of agent.

\noindent
\textbf{Decision Tree (DT) approach}: \quad The decision tree was applied to get a reference when cyber-attackers or cyber-defenders already know the best action to take as the role of each agent is defined by a DT.
In Figure~\ref{fig:graphs}, the $DT\_attacker1$ follows an action path to reach the goal of installing a custom spyware in PS. In the same time, $DT\_attacker2$ reaches the goal of exfiltrating data in DB and completing the action path by getting the flag. Then we added the defenders $DT\_defender1$ which has to detect malicious logs on WS and $DT\_defender2$ which has to use privilege account management or monitor the executed commands and arguments on DB. We observed the attackers to be unable to reach the ultimate goal.

\noindent
\textbf{Multi-Agent Reinforcement Learning (MARL) approach}: \quad Q-Learning~\cite{CWatkins1992} was applied with curriculum learning for first the attackers learn how to reach the ultimate attack goal before adding defenders.
In the Figure~\ref{fig:graphs}, $MARL\_attacker1$ and $MARL\_attacker2$ follow a same behavior ultimately refining the applied actions to the relevant ones to reach the ultimate goal. After several episodes, chosen action paths by the attackers tend to be as efficient as the DT paths. When adding the defenders $MARL\_defender1$ and $MARL\_defender2$, we verified the attackers to be less and less able to reach the ultimate goal.

\section{Conclusion and perspectives}

\noindent
We proposed a Dec-POMDP modeling of networked node likely to be attacked and defended by agents. This model aims to integrate scenarios. The implementation of this model led to a simulator whose some capabilities have been assessed through a MITRE ATT\&CK scenario. Using three approaches, we briefly checked how decision tree, random and reinforcement learning approaches can be applied to agent for comparison.
Willing to take advantage of this simulation approach to address realistic issues related to cyber-defenders, particularly in the AICA context, we identified the main limitations to overcome:
automating the integration of more realistic scenarios leveraging on a basis of common actions and properties so agents can explore and act as in seemingly similar to reality information systems;
establishing a way to use the benefits of results obtained with simulations for emulated or real systems while maintaining agent behaviors during deployment;
having more coordination between agents, such as several entry points or scenarios with needed communication to reach a goal\dots;
and introducing new constraints in actions (such as cost, execution duration, etc.).

\section*{References}

\bibliographystyle{IEEEtran}

\bibliography{local_references}

\begin{thebibliography}{10}
\providecommand{\url}[1]{#1}
\csname url@samestyle\endcsname
\providecommand{\newblock}{\relax}
\providecommand{\bibinfo}[2]{#2}
\providecommand{\BIBentrySTDinterwordspacing}{\spaceskip=0pt\relax}
\providecommand{\BIBentryALTinterwordstretchfactor}{4}
\providecommand{\BIBentryALTinterwordspacing}{\spaceskip=\fontdimen2\font plus
\BIBentryALTinterwordstretchfactor\fontdimen3\font minus \fontdimen4\font\relax}
\providecommand{\BIBforeignlanguage}[2]{{%
\expandafter\ifx\csname l@#1\endcsname\relax
\typeout{** WARNING: IEEEtran.bst: No hyphenation pattern has been}%
\typeout{** loaded for the language `#1'. Using the pattern for}%
\typeout{** the default language instead.}%
\else
\language=\csname l@#1\endcsname
\fi
#2}}
\providecommand{\BIBdecl}{\relax}
\BIBdecl

\bibitem{MITREATTACKWebiste}
``{MITRE ATTA\&CK},'' \url{https://attack.mitre.org/}, accessed: 2023-04-11.

\bibitem{CPhilips1998}
C.~Phillips and L.~P. Swiler, ``A graph-based system for network-vulnerability analysis,'' in \emph{Proceedings of the 1998 Workshop on New Security Paradigms}, 1998.

\bibitem{BKordy2010}
{Kordy, Barbara et. al.}, ``Foundations of attack--defense trees,'' in \emph{Formal Aspects of Security and Trust}, Berlin, Heidelberg, 2011, pp. 80--95.

\bibitem{MPetty2022}
{Petty, Mikel D et al.}, ``Modeling cyberattacks with extended petri nets,'' in \emph{Proc. of the 2022 ACM Southeast Conference}, 2022.

\bibitem{JBland2020}
{Bland, John A. et al.}, ``Machine learning cyberattack strategies with petri nets with players, strategies, and costs,'' in \emph{National Cyber Summit (NCS) Research Track}, Cham, 2020, pp. 232--247.

\bibitem{SYamaguchi2020}
\BIBentryALTinterwordspacing
S.~Yamaguchi, ``White-hat worm to fight malware and its evaluation by agent-oriented petri nets,'' \emph{Sensors}, vol.~20, no.~2, 2020. [Online]. Available: \url{https://www.mdpi.com/1424-8220/20/2/556}
\BIBentrySTDinterwordspacing

\bibitem{MPanfili2018}
{Panfili, Martina et. al.}, ``A game-theoretical approach to cyber-security of critical infrastructures based on multi-agent reinforcement learning,'' in \emph{26th Mediterranean Conf. on Control and Automation}, 2018.

\bibitem{AAttiah2018}
{Afraa, Attiah et al.}, ``A game theoretic approach to model cyber attack and defense strategies,'' in \emph{IEEE Conf. on Communications}, 2018, pp. 1--7.

\bibitem{CXiaolin2008}
{Xiaolin, Cui et. al.}, ``A markov game theory-based risk assessment model for network information system,'' in \emph{2008 Int. Conf. on Computer Science and Software Engineering}, vol.~3, 2008, pp. 1057--1061.

\bibitem{beynier2010}
\BIBentryALTinterwordspacing
{Beynier, Aur{\'e}lie et al.}, ``{DEC-MDP / DEC-POMDP},'' in \emph{{Markov Decision Processes in Artificial Intelligence}}, 2010, pp. 277--313. [Online]. Available: \url{https://hal.science/hal-00969197}
\BIBentrySTDinterwordspacing

\bibitem{jk2020}
{Terry, J. K et al}, ``Pettingzoo: Gym for multi-agent reinforcement learning,'' 2020.

\bibitem{bernstein2013}
\BIBentryALTinterwordspacing
{Daniel S. Bernstein et al.}, ``The complexity of decentralized control of markov decision processes,'' \emph{CoRR}, vol. abs/1301.3836, 2013. [Online]. Available: \url{http://arxiv.org/abs/1301.3836}
\BIBentrySTDinterwordspacing

\bibitem{theron_autonomous_2021}
P.~Theron, N.~Evans, M.~Drasar, and A.~Guarino, ``{Autonomous} {Intelligent} {Cyber} {Defence} {Agent} {Prototype} 2021 - {Project} {Report},'' Dec. 2021.

\bibitem{OliehoekA16}
\BIBentryALTinterwordspacing
F.~A. Oliehoek and C.~Amato, \emph{A Concise Introduction to Decentralized POMDPs}, ser. Springer Briefs in Intelligent Systems.\hskip 1em plus 0.5em minus 0.4em\relax Springer, 2016. [Online]. Available: \url{https://doi.org/10.1007/978-3-319-28929-8}
\BIBentrySTDinterwordspacing

\bibitem{GFalco2018}
{Falco, Gregory et al.}, ``A master attack methodology for an ai-based automated attack planner for smart cities,'' \emph{IEEE Access}, vol.~6, pp. 48\,360--48\,373, 2018.

\bibitem{DGrunewald2011}
{Grunewald, Dennis et. al.}, ``Agent-based network security simulation,'' vol.~2, 01 2011, pp. 1325--1326.

\bibitem{IKotenko2007}
I.~Kotenko, ``Multi-agent modelling and simulation of cyber-attacks and cyber-defense for homeland security,'' in \emph{2007 4th IEEE Workshop on Intelligent Data Acquisition and Advanced Computing Systems: Technology and Applications}, 2007, pp. 614--619.

\bibitem{Varga2010}
\BIBentryALTinterwordspacing
A.~Varga, \emph{OMNeT++}.\hskip 1em plus 0.5em minus 0.4em\relax Berlin, Heidelberg: Springer Berlin Heidelberg, 2010, pp. 35--59. [Online]. Available: \url{https://doi.org/10.1007/978-3-642-12331-3_3}
\BIBentrySTDinterwordspacing

\bibitem{drasar_session-level_2020}
{Drasar, Martin et. al.}, ``Session-level adversary intent-driven cyberattack simulator,'' in \emph{2020 {IEEE}/{ACM} 24th Int. Symp. on Distributed Simulation and Real Time Applications}, 2020.

\bibitem{cyberbattlesim}
\BIBentryALTinterwordspacing
M.~D.~R. Team., ``Cyberbattlesim,'' 2021, created by {C. Seifert et. al.} [Online]. Available: \url{https://github.com/microsoft/cyberbattlesim}
\BIBentrySTDinterwordspacing

\bibitem{MCASWebsite}
``Multi cyber agent simulator,'' \url{https://github.com/julien6/MCAS}, accessed: 2023-04-17.

\bibitem{CWatkins1992}
C.~Watkins and P.~Dayan, ``Technical note: Q-learning,'' \emph{Machine Learning}, vol.~8, pp. 279--292, 05 1992.

\end{thebibliography}

\end{document}